\pdfoutput=1

\documentclass[11pt]{article}

\usepackage[preprint]{acl}
\usepackage{times}
\usepackage{latexsym}
\usepackage[T1]{fontenc}

\usepackage[utf8]{inputenc}

\usepackage{microtype}

\usepackage{inconsolata}

\usepackage{graphicx}

\newcommand{\cicero}{\abr{Cicero}\xspace }
\newcommand{\amr}[1]{\texttt{#1}}

\usepackage{times}
\usepackage{latexsym}
\usepackage{enumitem}
\usepackage{tikz}
\setlist{nosep}
\usepackage{tabularx}
\usepackage{microtype}
\usepackage{graphicx}
\usepackage{booktabs} 
\usepackage{amsmath}
\usepackage{amsfonts}
\usepackage{subcaption}
\usepackage[all]{nowidow}
\definecolor{grayish}{rgb}{0.95, 0.95, 0.96}

\usepackage{xcolor}
\usepackage{hyperref}
\usepackage{inconsolata}
\newcommand{\email}[1]{\textcolor{blue}{\small\texttt{\href{mailto:#1}{#1}}}}
\newif\ifcomment\commentfalse

\definecolor{fra}{RGB}{65, 105, 225}
\definecolor{ita}{RGB}{34, 139, 34}
\definecolor{ger}{RGB}{160, 138, 117}
\definecolor{aus}{RGB}{255, 0, 0}
\definecolor{tur}{RGB}{185, 166, 28}

\usepackage{xspace}

\title{Should I Trust You? Detecting Deception in Negotiations using Counterfactual RL}

\author{Wichayaporn Wongkamjan$^{1}$ \hspace{0.5cm} Yanze Wang$^{4}$ \hspace{0.5cm} Feng Gu$^{1}$ \hspace{0.5cm} Denis Peskoff$^{2}$ \\
{\bf Jonathan K. Kummerfeld}$^{3}$ \hspace{0.5cm} {\bf Jonathan May}$^{4}$ \hspace{0.5cm} {\bf Jordan Lee Boyd-Graber}$^{1}$ \\
$^{1}$\small{Department of Computer Science, University of Maryland} \hspace{0.5cm}
$^{2}$\small{Northwestern University} \\
$^{3}$\small{School of Computer Science, University of Sydney} \\
$^{4}$\small{Information Sciences Institute, University of Southern California} \\
\email{wwongkam@umd.edu} \hspace{0.2cm} \email{yanzewan@isi.edu} \hspace{0.2cm} \email{fgu1@umd.edu}\\ 
\email{jonathan.kummerfeld@sydney.edu.au} \hspace{0.2cm} 
\email{jonmay@isi.edu} \hspace{0.2cm} 
\email{jbg@umiacs.umd.edu} 
}

\usepackage[a-1b]{pdfx}

\usepackage{framed}
\usepackage{mdwlist}
\usepackage{siunitx}
\usepackage{latexsym}
\usepackage{colortbl}
\usepackage{xcolor}
\usepackage{nicefrac}
\usepackage{booktabs}
\usepackage{fnpct}
\usepackage{amsfonts}
\usepackage[T1]{fontenc}
\usepackage{bold-extra}
\usepackage{amsmath}
\usepackage{amssymb}
\usepackage{bm}
\usepackage{graphicx}
\usepackage{mathtools}
\usepackage{microtype}
\usepackage{multirow}
\usepackage{multicol}
\usepackage{xpatch}
\usepackage{latexsym,comment}
\usepackage[normalem]{ulem}

\newcommand*{\missingreference}{{\Huge \colorbox{red}{?reference?}}}
\newcommand*{\missingcitation}{{\Huge \colorbox{red}{?citation?}}}

\makeatletter
\xpatchcmd{\@setref}{\bfseries}{\missingreference}{}{}
\def\@citex[#1]#2{\leavevmode
    \let\@citea\@empty
    \@cite{\@for\@citeb:=#2\do
        {\@citea\def\@citea{,\penalty\@m\ }%
            \edef\@citeb{\expandafter\@firstofone\@citeb\@empty}%
            \if@filesw\immediate\write\@auxout{\string\citation{\@citeb}}\fi
            \@ifundefined{b@\@citeb}{\hbox{\reset@font\missingcitation}%
                \G@refundefinedtrue
                \@latex@warning
                {Citation `\@citeb' on page \thepage \space undefined}}%
            {\@cite@ofmt{\csname b@\@citeb\endcsname}}}}{#1}}
\makeatother

\newcommand{\gem}[1]{\mbox{\textsc{gem}}}
\newcommand{\abr}[1]{\textsc{#1}}

\newcommand{\hidetext}[1]{}
\newcommand{\ignore}[1]{}

\ifcomment
    \newcommand{\pinaforecomment}[3]{\colorbox{#1}{\parbox{.8\linewidth}{#2: #3}}}
    \newcommand{\commentboxinbracket}[3]{{\color{#1} [#3]$_\text{#2}$}}

    \newcommand{\prtodo}[1]{\pinaforecomment{lightblue}{pr}{#1}}
    \newcommand{\prtodoi}[1]{\pinaforecomment{lightblue}{pr}{#1}}
\else
    \newcommand{\pinaforecomment}[3]{}
    \newcommand{\prtodo}[1]{}
    \newcommand{\prtodoi}[1]{}
    \newcommand{\commentboxinbracket}[3]{}
\fi

\newcommand{\jbgcomment}[1]{\pinaforecomment{red}{JBG}{#1}}

\newcommand{\wwcomment}[1]{\pinaforecomment{yellow}{Joy}{#1}}
\newcommand{\jmcomment}[1]{\pinaforecomment{magenta!20}{Jon}{#1}}
\newcommand{\fgcomment}[1]{\pinaforecomment{green!20}{Feng}{#1}}
\newcommand{\yzcomment}[1]{\pinaforecomment{green}{yanze}{#1}}

\newcommand{\smallurl}[1]{ \begin{tiny}\url{#1}\end{tiny}}

\definecolor{lightblue}{HTML}{3cc7ea}
\definecolor{CUgold}{HTML}{CFB87C}
\definecolor{grey}{rgb}{0.95,0.95,0.95}
\definecolor{ceil}{rgb}{0.57, 0.63, 0.81}
\definecolor{UMDred}{HTML}{ed1c24}
\definecolor{UMDyellow}{HTML}{ffc20e}

\begin{document}
\maketitle

\begin{abstract}

An increasingly common socio-technical problem is people being taken in by offers that sound ``too good to be true'', where persuasion and trust shape decision-making.
This paper investigates how \abr{ai} can help detect these deceptive scenarios.
We analyze how humans strategically deceive each other in \textit{Diplomacy}, a board game that requires both natural language communication and strategic reasoning.
This requires extracting logical forms representing proposals---agreements that players suggest during communication---and computing their relative rewards using agents' value functions.
Combined with text-based features, this can improve our deception detection.
Our method detects human deception with a high precision when compared to a Large Language Model approach that flags many true messages as deceptive.
Future human-\abr{ai} interaction tools can build on our methods for
deception detection by triggering \textit{friction} to give users a
chance of interrogating suspicious proposals.\footnote{Code is available at \href{https://github.com/ALLAN-DIP/diplomacy_cicero/tree/deception_friction_value/fairdiplomacy_external/friction}{CTRL-D github repository}}
%
%
\end{abstract}

\section{Friction in \abr{ai} systems}
\begin{figure}[ht]
    \centering
    \includegraphics[width=0.45\textwidth]{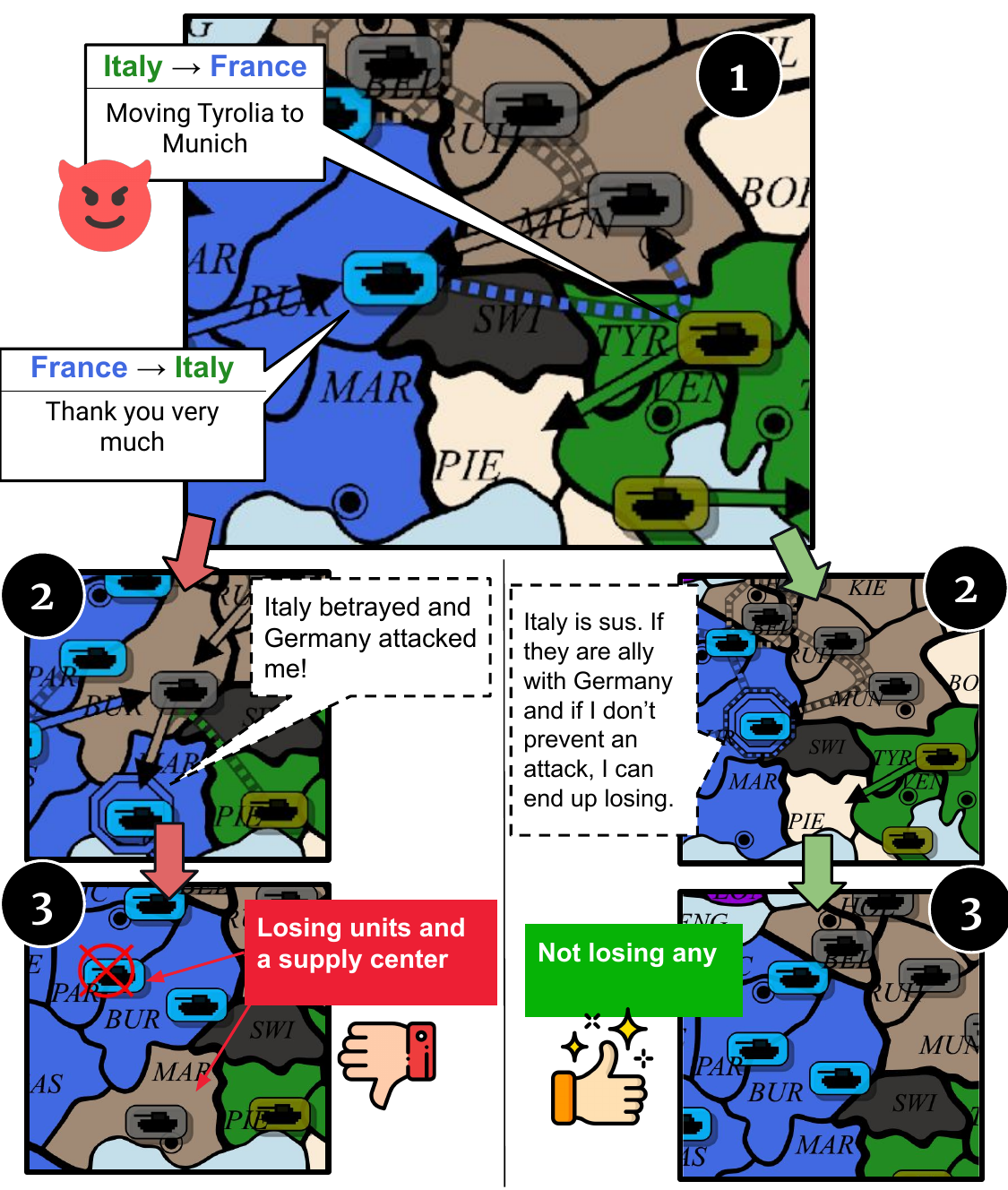}
    \caption{Detecting deception is crucial in mixed cooperative-competitive environments. \textbf{(Left)} \textcolor{fra}{France} believed the lie that \textcolor{ita}{Italy} will move their army in Tyrolia to Munich, losing Burgundy and subsequently Marseilles to \textcolor{ger}{Germany}. \textbf{(Right)} If \textcolor{fra}{France} had detected the deception, they could have successfully defended Burgundy and avoided disbanding one army.}
    \label{fig:intro_example}
\end{figure}


\begin{figure*}[h]
    \centering
    \includegraphics[width=\textwidth]{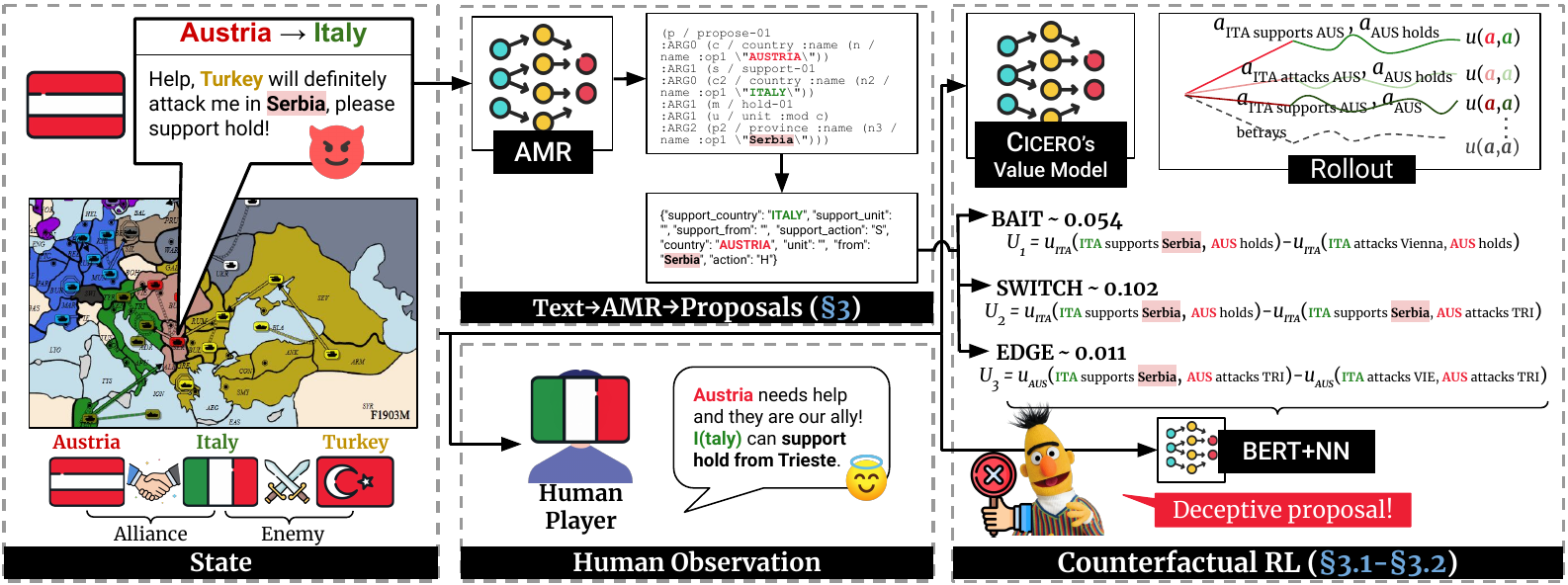}
    \caption{An overview of our approach to detect deceptive proposals, requiring a recipient (Human Player) to follow a proposed action. \textbf{(Left)} A state of this Diplomacy game is (1) \textcolor{aus}{Austria} and \textcolor{ita}{Italy} have an alliance (2) while \textcolor{tur}{Turkey} and \textcolor{ita}{Italy} have been clashing for several turns. \textcolor{aus}{Austria} realizes that they are in a weak spot and need a quick escape, so they reach out to \textcolor{ita}{Italy}. It is a deceptive proposal so that \textcolor{aus}{Austria} can get to Trieste. \textbf{(Bottom Middle)} A human player can be biased towards their own ally (\textcolor{aus}{Austria}) and use their fast-thinking system to instinctively help. \textbf{(Top Middle)}  For an alternative perspective, our approach converts natural language to proposals using \abr{amr}. \textbf{(Right)} Then, we leverage the RL value function from \cicero to estimate three aspects of deception---Bait, Switch and Edge---from counterfactual actions of \textcolor{aus}{Austria} and \textcolor{ita}{Italy}. 
    Passing the dialogue alongside these values to a classifier decides whether \textcolor{aus}{Austria}'s proposal is \textbf{deceptive}.}
    \label{fig:overview}
\end{figure*}

Deception in natural language is a fundamental aspect of human communication, often employed as a strategic tool to mislead others through misrepresentation, omission, exaggeration, or counterfactual reasoning~\citep{bok2011lying}. 
From casual social interactions to high-stakes negotiations, deception influences trust, decision-making, and cooperation, making it a subject of extensive study in psychology, linguistics, and philosophy, manifesting real-world challenges such as fake news on
social media ~\citep{bade-etal-2024-social}, misinformation ~\citep{panda-levitan-2022-improving} and adversarial communication in strategic
games ~\citep{bernard-mickus-2023-many}. As artificial intelligence systems increasingly engage in human-like communication, they not only inherit but also amplify deceptive strategies, sometimes unintentionally.
In \abr{ai}-generated text, deception can emerge as a byproduct of
optimization objectives, particularly in \abr{rlhf} where
agents maximize utility in multi-agent settings, sometimes at the
expense of honesty~\cite{wen2025language}.
This phenomenon has garnered significant attention across various
domains, as \abr{ai} deception is not confined to theoretical
constructs but manifests in real-world challenges, e.g. hallucination in reasoning tasks ~\citep{grover-etal-2024-navigating}.


Prior research underscores that \abr{ai}-generated deceptive
communication can be difficult to detect and may lead to unintended
consequences when deployed in practical applications~\citep{park2024ai,sarkadi2024deceptive}.
Deceptive AI-generated text can erode trust in digital communication,
amplify misinformation, and facilitate large-scale manipulation in
political, financial, and social domains~\citep{solaiman2019release,
  weidinger2022taxonomy}.
Furthermore, the scalability of AI models accelerates the production of deceptive content and dissemination of making manual detection impractical.
%
To address these risks, robust mitigation strategies are necessary,
including adversarial training~\citep{perez2022red}, explain-ability
techniques to enhance AI transparency~\citep{danilevsky2020explainability}, and real-time detection methods
leveraging linguistic and behavioral cues~\citep{vosoughi2018spread}.

We test our detection strategies within the environment of  \textit{Diplomacy}, a game rich in negotiation, cooperation, and betrayal expressed through natural language. The most intriguing moments of the game arise when two players negotiate to cooperate in pursuit of their respective goals. While such agreements usually yield mutual benefits, this is not always the case---some negotiated arrangements are the result of deception, omission, or straight-up lies on the part of one player. Skilled players combat such behavior by developing the ability to recognize when an offer \textit{sounds too good to be true},  whereas typical players struggle to recognize such situations~\cite{peskov2020takes,wongkamjan-etal-2024-victories,gu-etal-2025-personalized}. Our work explores this area to raise awareness among human players when they encounter deception embedded in negotiations. 

We leverage the value function---an \abr{RL} function  that estimates how good a move and a position is---learned by \cicero~\citep{bakhtin2023mastering}, the strongest \abr{ai} agent trained to play Diplomacy at a human level, to evaluate whether a proposal is ``\textit{too good to be true.}'' 
Our contributions are as follows:\\
\begin{enumerate}
    \item With Theory-of-Mind-influenced deception, we identify negotiations in natural language via formal logical modeling and detect potential deceptive offers in negotiations using the \cicero \abr{rl} value function to generate counterfactual explanations.
    \item We train a BERT-based~\cite{devlin-etal-2019-bert} classifier to predict deception using \abr{rl} values and message embeddings.
    \item Our classifier is more accurate than a fine-tuned Llama3 in human lie prediction and detecting partially-deceptive negotiations. 
\end{enumerate}

\section{Deception in the Wild}

Real-world deception manifests in various forms, such as \textit{scams} and \textit{phishing} attacks, where perpetrators exploit \textbf{trust} to manipulate victims into believing in the possibility of good fortune, even if it is unlikely~\citep{button2014online,muscanell2014weapons,hanoch2021scams}. These deceptive tactics often rely on persuasive language. If victims fall for these \textbf{too good to be true} claims, they become targets and may comply with the perpetrators' requests---for example, disclosing sensitive information or making financial investments under false pretenses---ultimately resulting in monetary \textit{loss} or data breaches~\citep{burnes2017prevalence,coluccia2020online}. Those scammers would \textit{gain value} through data breaches or simply by acquiring cash.

Detecting deception remains a persistent challenge, especially when it is needed for real-world problems.
The lack of realistic, thorough deception training data precludes supervised \abr{ai} approaches. 
Deception in a limited space like a strategic game, e.g., Diplomacy, where nuanced persuasion and deception is required for winning, is more tractable to evaluate.
A bounded example would allow us to measure the ability of an \abr{ai} to improve in deception detection.

%
\jmcomment{Why `Thus'? It doesn't follow from what comes before. I think it would be fine to just describe how it's tricky to deceive and tricky to detect deception. However if we're dealing with detection then we should focus on that difficulty, not the difficulty in creation.} 
\wwcomment{I rewrote this section to deliver better}

\subsection{One Gains, One Loses}

Deception has been studied in games that rely on trust, negotiation, and strategic misrepresentation, such as Werewolf~\citep{chittaranjan2010you,hancock2017towards,girlea2017deception}, Poker~\citep{lee2013deception, palomaki2016machiavelli}, and Diplomacy~\citep{niculae-etal-2015-linguistic,kramar2022negotiation,baldwin2025strat,kulkarni2025dynamic}. Diplomacy is a complex interplay of strategy, high-level cooperation, and subtle betrayal. The game is set on an European map, highlighting key territorial cities known as supply centers. Each of the seven players controls a country and moves units on the map, with the objective of capturing more than half of the supply centers (18 out of 34) to achieve victory. For each turn, players communicate one-to-one and then simultaneously reveal their orders for each units.

Deception plays a crucial role in gaining supply centers and, ultimately, securing a win. Cliques of players agreeing to coordinate to gain advantages over others must operate in secrecy. \jmcomment{I thought more explanation was needed but I am not yet satisfied with my amendment \wwcomment{I will come back to this!}} 
Deception must be undetected to be successful. If the player fails to recognize deception, they risk losing supply centers and may lose the game (Figure \ref{fig:intro_example}). If a player's deception succeeds, they may gain supply centers. The challenge lies in quantifying the benefits of deception and the losses of those who are deceived. Given the significance of supply centers as a sparse scoring mechanism, we see an opportunity to integrate reinforcement learning~\citep[\abr{rl},][]{zinkevich2007regret,brown2019deep} into the analysis.

\abr{rl} has been extensively used to train \abr{ai} agents in optimizing decision-making that maximize a reward. \abr{rl}-based \abr{ai} has been applied to Diplomacy \cite{NEURIPS2019_84b20b1f,anthony2020learning,gray2021humanlevel,bakhtin2021nopress}, with a recent model, 
\cicero~\citep{meta2022human}, achieving competitive human-level play. This paper uses a value model from \cicero{} to estimate the expected future rewards of a proposed move, enabling the detection of proposals that are likely deceptive (Figure~\ref{fig:overview}).

\section{Counterfactual \abr{rl} against Deception}
\label{sec:overall_method}

\jmcomment{This section is way too insider-y. I think all that matters here is we want to know if a proposed message is deceptive. I think it's fine to just state that and move on, and leave the parse-and-find-propose-01 bit out or maybe in the low level details of the code. \wwcomment{removed those amr details part}}
We look for deception\jmcomment{terminology needs to be introduced now \wwcomment{I removed system2}} in the text messages between pairs of players.\jmcomment{unclear why text and private dialogues are highlighted here unless we're distinguishing them from smth else \wwcomment{fixed}}
Each player controls multiple units in this game, so we restrict a pool of messages where a player explicitly requests another player to issue a specific order (e.g. Austria asks Italy to support in Serbia, Figure~\ref{fig:overview}). \fgcomment{agree with Jono; not clear} With this, we parse messages in natural language to Abstract Meaning Representation~\citep[\abr{amr},][]{banarescu-etal-2013-abstract}. 

For any message to a player, we want to raise awareness if the proposal is \textit{potentially deceptive}. We leverage a well-trained  value function, a part of \cicero~\citep{meta2022human}, to estimate how likely a proposal is deceptive. This section we discusses our method, \textbf{C}oun\textbf{T}erfactual \textbf{\abr{rl}} against \textbf{D}eception (CTRL-D), which has two main components: 1) Counterfactual RL and 2) formulations to capture potential deceptive proposals.

\subsection{Counterfactual \abr{rl}}
\label{sec:cfrl}
Player \( i \) needs to pick an action \( a_i \) given a board state \( s \). However, moves in Diplomacy do not happen in isolation---all actions of \emph{other} players \( a_{-i} \) happen simultaneously, so \cicero{} uses a function \( u_i(a_i, a_{-i},s) \) that represents estimated future rewards that player \( i \) will receive if actions \( a_i \) and \( a_{-i} \) are played in a state \( s \). Thus, a high value represents a ``\textit{better}'' move based on learned strategies.

While a review of \cicero is outside the scope of this paper, its value function allows our work to compute counterfactual one-step actions to estimate potential deceptive proposals from another player \( j \), where each proposal is about action \( a_i \) and \( a_j \). Equipped with text-to-proposals and the \abr{rl} value function, we are ready to detect deception. 

\subsection{Deceptive Proposals}
\label{sec:dec_prop}

To estimate how likely a proposal is deceptive, we introduce three \textit{deceptive signs} that account for different aspects of deception.
First, we can measure whether a victim would get a higher reward if the proposal was not a deception (i.e., is the fake proposal from the deceiver appealing?). Second, we can measure whether a victim would get a lower reward if they believe the deception.
Third, we can measure whether a deceiver would increase their future reward by deceiving the victim. We call these three measures: Bait, Switch and Edge\footnote{These terms come from popular culture terms around scams: a deceiver offers ``\textit{bait}'' to attract the victim who suffers from the ``\textit{switch}'', leading the deceiver to profit, their ``\textit{edge}'' in the scam}.
In this section, we highlight deceptive values through three hypotheses.

We define a proposal \( p_{j\to i} \) when player \( j \) proposes actions to player \( i \). A proposal \( p_{j\to i} \) consists of an action \( \hat{a}_i \) that player \( j \) wants player \( i \) to play and an action \( \hat{a}_j \) that player \( j \) promises to make. In Diplomacy, an action is a tuple of unit orders, e.g. \jmcomment{Do you really mean `moves' and not more general orders? Are supports and holds involved? All the examples are actual moves but i'm unclear if that restriction is intended.}
\wwcomment{to make it more clear, I will be using unit orders.}
\begin{itemize}
    \item \colorbox{yellow}{an army in Berlin moves to Kiel},
    \item \colorbox{pink}{an army in Munich moves to Ruhr} and
    \item \colorbox{lime}{a fleet in Kiel moves to Holland}
\end{itemize} 
where these can represent in logical forms as \amr{(`\colorbox{yellow}{A BER - KIE}',`\colorbox{pink}{A MUN - RUH}',
`\colorbox{lime}{F KIE - HOL}')}.
Player \( j \) can propose an action to player \( i \) with multiple unit orders, denoted \( \hat{a}_i = (\hat{a}_{i,1} ,\hat{a}_{i,2}, \dots,  \hat{a}_{i,n}) \), where \( n \) is a number of player \( i \)'s units. 
\fgcomment{for the paragraphs below, should we use player/deceiver+victim interchangeably?}
We estimate how likely a proposal \( p_{j\to i} \) is deceptive by following three hypotheses when it is `\textit{too good to be true}''.
\subsubsection{Bait}
A victim \textit{perceives a greater reward} if they \textit{alter} a decision to \textit{follow} the deceiver's proposal and the deceiver does not actually deceive, but rather follows the plan. \fgcomment{not clear and grammar issue}
Assume player \( i \) has a plan \( a_i\), and player \( j \) proposes \( \hat{a}_i \) and \( \hat{a}_j \). From the perspective of player \( i \), they decide to play \( \hat{a}_i \) because they perceive that the estimated future rewards will increase by: 

\begin{equation}
    \begin{aligned}
        U_1 = u_i(\hat{a}_i, \hat{a}_j) - u_i(a_i, \hat{a}_j).
    \end{aligned}
    \label{eq:deception_hypo3}
\end{equation}
\subsubsection{Switch}
A victim will \textit{receive a lower} reward if they \textit{follow} the deceiver's request and if the deceiver betrays the victim.
 Player~\( j \) proposes actions \( \hat{a}_i \) and \( \hat{a}_j \) to player \( i \) where player~\( j \) has alternative plan \( a_j \) to instantly stab or take advantage of player~\( i \). The estimated future rewards of player~\( i \) will decrease if player~\( j \) betrays player~\( i \). We leverage \cicero's \abr{rl} value function \( u_i \) (Section~\ref{sec:cfrl}) to formulate the first hypothesis:

\begin{equation}
    \begin{aligned}
        U_2 = u_i(\hat{a}_i, \hat{a}_j) - u_i(\hat{a}_i, a_j)
    \end{aligned}
    \label{eq:deception_hypo1}
\end{equation}
where \( a_j \) is an alternative action that player \(j\) might play instead of following the proposed move
\mbox{\(a_j\neq\hat{a}_j\)}. 

\subsubsection{Edge}
A deceiver will \textit{receive a better reward} when a victim \textit{follows} their proposal. Given the deceiver~\( j \)'s plan \( a_j \)  and the victim's plan \( a_i\), if player~\( j \) proposes a suboptimal \( \hat{a}_i \) to player~\( i \) and player~\( i \) falls for it. The estimated future rewards for player~\( j \) can increase:  
\begin{equation}
    \begin{aligned}
        U_3 = u_j(\hat{a}_i, a_j) - u_j(a_i, a_j).
    \end{aligned}
    \label{eq:deception_hypo2}
\end{equation}

In short, the three hypotheses for deceptive proposals assume the victim loses, the deceiver gains, and the victim follows the proposal~(Counterfactual RL, Figure \ref{fig:overview}). For player~\(i\)'s plan \( a_i\) in Equation \ref{eq:deception_hypo3} and Equation \ref{eq:deception_hypo2}, we define \( a_i\)  as the optimal action from player~\(i\)'s perspective, thus sampling an action \( a_i \sim \pi_i\) where \(\pi_i\) is \cicero{}'s policy.

In the final step, we train a neural network classifier, three-layer linear layers with a Binary Cross Entropy loss by inputting two main features, \textit{text embedding} and \textit{numeric features} (shape=\amr{hidden\_size + 3}).
\begin{itemize}
    \item We retrieve \textit{text embedding} by passing a text input to BERT, where we use a [CLS] token representation in the BERT output as a summarization of sentence (shape=\amr{hidden\_size})
    \item We directly use deceptive values $U_1$, $U_2$, $U_3$ as \textit{numeric features} (shape=\amr{3}). 
    \item We concatenate \textit{text embedding} and \textit{numeric features} and further pass them to the classifier and output a float value ranging from $[0,1]$ to predict deception probability.
\end{itemize}


We train only ten epochs with a small training data sampled from \citet{peskov2020takes}. In the next section, we discuss datasets that we use to train the classifier and to test our approach against an LLM baseline.

\section{Recall-Oriented Lie Detection for Friction}
\label{sec:setup}
This section explores deception detection and its role in creating strategic friction, i.e., deliberate decision-making in human-\abr{ai} interactions. Section \ref{sec:denis_data} analyzes human-only Diplomacy games~\citep{peskov2020takes} to categorize deceptive messages and select messages for training and evaluation. Section \ref{sec:friction} extends this analysis to a larger dataset within human-\abr{ai} settings, testing whether our framework can introduce friction against deceptive proposals. Our goal is not to optimize $F_1$-Score but rather to flag \textit{possible} deception for users, introducing friction to help them detect deception. In other words, our goal is to examine these cases.
%
\subsection{Alignment to Human Lies}
\label{sec:denis_data}
To understand human deception, we use the twelve Diplomacy games\footnote{Github: \href{https://github.com/DenisPeskoff/2020_acl_diplomacy}{It Takes Two to Lie: One to Lie, and One to Listen}} annotated by~\citet{peskov2020takes} containing human strategy through orders and communication from private messages. The dataset contains annotations from players at per-message granularity, indicating whether or not the content of their message contained a lie (Table~\ref{fig:lie_annotation}). \jmcomment{unclear if the rest of this graf is needed \wwcomment{I removed it to save space :)}} 
\begin{table}[t]
    \begin{tabular}{p{1.2cm}p{5.4cm}}
    \hline
    \textbf{Sender} & \textbf{Message}   \\ 
    \hline
     \rowcolor{grayish} Russia (Lie) & \small{I think I will move Moscow into War, with Sil supporting, where I could go for Austria the following turn.} \\ 
    \rowcolor{grayish}  
    Russia (Lie) & \small{Also I will move Ukraine to gal. Could you support me there?} \\
     Turkey (Truth) & \small{yeah I dont mind support} \\
  \hline
    \end{tabular}
    \caption{An example of a lie annotation from a human player in \citet{peskov2020takes} dataset.}
    \label{fig:lie_annotation}
\end{table}
\begin{table}[t]
\centering
\begin{tabular}{lr}
\hline
 Categories&Total\\ 
\hline
Any messages& 17,289\\
\hline
Any lies&842 \\
\hspace{3mm} Other &459\\
\rowcolor{lime} \hspace{3mm} \textbf{Deceptive Moves}&\textbf{286}\\
 \hspace{3mm} Feigning Trust/Loyalty&28 \\
 \hspace{3mm} False Excuse&27\\
 \hspace{3mm} Withholding Information&24\\
\hline
\end{tabular}
\caption{We categorize lie messages in \citet{peskov2020takes} data set, in which \textbf{Deceptive Moves} is the closest to our interest. Though messages with this type do not appear often, they are useful for our CTRL-D to get deceptive signs.}
\label{tab:denis_lies}
\end{table}
To best select training and evaluation data, first we breakdown lie messages---natural language texts generated by humans---in \citeauthor{peskov2020takes}'s dataset into categories (Table~\ref{tab:denis_lies}), where the category that is closest to our interest is \textbf{Deceptive Moves}.\footnote{For more category details, see Appendix~\ref{sec:denis_lie_cat}} \mbox{Deceptive} moves are rare, constituting fewer than 1.7\% of all messages; therefore, we sample data for training and evaluation:
\begin{itemize}
    \item \textbf{For training data,} we focus on messages 
    with human-annotated negotiations---logical form of negotiations annotated by experts---specifically for player \(i\) or player \(j\). 
    In total, there are 344 messages with human-annotated negotiations containing fifty-nine lies and twenty-eight proposals. We sample additional 1,500 messages, though without human annotation, we retrieve logical forms of negotiations as discussed in Section~\ref{sec:overall_method}. \jmcomment{This seems like an engineering detail regarding a system that hasn't been explained. It's usually a bad idea to put an actual vector of numbers in the text of a paper. On the whole I'm unclear what this section is trying to communicate.\wwcomment{simplify this too engineering step and try to connect each paragraph better. I will improve this more}}
    \item \textbf{For evaluation,} we sample 1,000 messages containing eighty lies from the rest of dataset without any further selection.
\end{itemize}


\subsection{Friction for Humans} 
\label{sec:friction}
While the data from \citet{peskov2020takes} confirms our approach: \textbf{CTRL-D} has a desired recall, it is small. Thus, we next test generalization on Meta's data set curated from \amr{webdiplomacy.net}\footnote{\href{https://ai.meta.com/research/request-for-proposal/towards-human-AI-cooperation/}{AI@Meta: Towards Human-AI Cooperation RFP}}, which contains 40,000 games, 13 million natural language interactions from humans and \cicero players. Although these data lack thorough deception annotation, we can ex post facto validate precision through human verification.

\section{Baselines for Deception}
\label{sec:llama}

Deception is not only about the offer on the table: \citet{niculae-etal-2015-linguistic} show language changes before a betrayal occurs in Diplomacy, and \citet{lai2020chicago} demonstrate that this also holds true even in online reviews. These findings motivate the use of linguistic signals alone to detect deception, independent of game mechanics or strategic models. Thus, we compare our approach to language only baselines. Specifically, we implement an LLM-based baseline, using \amr{LLaMA\,3.1-8B-Instruct}~\citep{llama3modelcard} as our primary model, to detect and mitigate deception in Diplomacy negotiations. Our objective is to determine whether a large language model can identify suspicious messages without relying on reinforcement learning (\abr{rl}). We adopt two complementary criteria for evaluating deception: \emph{Direct Judgment} and \emph{Alignment Judgment}. Detailed prompts are provided in Figure~\ref{fig:prompt_llama}.
\newpage
\paragraph{Direct Judgment}
We prompt the baseline to decide whether a negotiation message is deceptive by examining its textual content together with relevant contextual information. For each negotiation round, we provide:
\begin{enumerate}
    \item \textbf{Current board state:} A summary of player's positions and units (e.g., which territories are occupied by which units).
    \item \textbf{Negotiation message:} The specific proposal made by player~\(j\) to player~\(i\).
    \item \textbf{Predicted Orders \(\tilde{a}_j\):} The orders \cicero would issue for player~\(j\) under an RL policy, indicating how \cicero believes player~\(j\) will actually move.
\end{enumerate}
Given this information, the baseline determines whether the message from player~\(j\) is deceptive, returning a binary label. We do not impose a strict threshold at this stage; instead, we rely on the model’s ability to integrate textual cues and board-state context.

\paragraph{Alignment Judgment}
We propose \emph{Alignment Judgment} which measures consistency between \cicero's predicted orders and the orders proposed in the negotiation message. Since \cicero's strategy model is trained to be near optimal, it is reliable and we see this as an opportunity to provide predicted orders as additional information.

For each player~\(j\), \cicero predicts orders~\(\tilde{a}_j\) using an RL policy. These predicted orders are compared to the actions stated in the negotiation message. If player~\(j\) proposes \(p_{j \to i}\) with actions \( \hat{a}_i\) and \( \hat{a}_j\), but \cicero predicts alternative actions~\(\tilde{a}_j\) where \(\tilde{a}_j\neq\hat{a}_j\), this inconsistency may indicate deception. We define two sets: the number of misaligned orders \( O_m = \sum_k \boldsymbol{1}[\tilde{a}_{j,k} \neq \hat{a}_{j,k}] \), and the number of aligned orders \( O_a = \sum_k \boldsymbol{1}[\tilde{a}_{j,k} = \hat{a}_{j,k}] \), where \(k\) is the number of the units of player \(j\).
If \( O_m \geq O_a \), the message is classified as deceptive and the baseline triggers a \emph{risk alert}, informing player~\(i\) of the misaligned orders. This prompts players to re-evaluate their decisions before finalizing their moves.

\section{Results and Analysis}
\label{sec:results}
\begin{table*}[h]
    \centering
    \begin{tabular}{lccc}
        \hline
        \textbf{Model} & \textbf{Precision} & \textbf{Recall} & \textbf{F1-Score} \\
        \hline
        LLM baseline using Direct Judgment & 0.095 & \textbf{0.551} &0.161 \\
        LLM baseline using Alignment Judgment & 0.147 & 0.065 & 0.090  \\
        CTRL-D \textbf{(ours)} & \textbf{0.950} & 0.238 & \textbf{0.380} \\
        \hspace{3mm}  CTRL-D with annotated logical forms  & \textbf{0.960} & 0.300 & \textbf{0.457} \\
        \hspace{3mm}  CTRL-D with only proposals & 0.868 & \textbf{0.413} & \textbf{0.560} \\
        Context LSTM + Power 
         & 0.263 & 0.171 & 0.207 \\ 
         Human suspected lies & 0.252 & 0.203 & 0.225 \\ 
        \hline
    \end{tabular}
    \caption{While \textbf{LLM baseline Direct Judgment} detects deception on actual human lies with a high recall, its precision is very low. \textbf{LLM baseline using Alignment Judgment} and \citet{peskov2020takes} \textbf{LSTM} shows problems in detection with poor precision and recall. Players from \citet{peskov2020takes} struggle to recognize lies, highlighting that our \textbf{CTRL-D} has critically high precision and the best overall between precision and recall.}
    \label{tab:1K_results}
\end{table*}

This section compares the approach (Section~\ref{sec:overall_method}) ---a classifier trained by using \textit{deceptive values} from counterfactual RL and text embeddings from BERT to train a three-layer linear neural network---to an LLM baseline (Section~\ref{sec:llama}). We evaluate these using two Diplomacy datasets; the first is~\citet{peskov2020takes} with human lie annotations\fgcomment{what is the other one?}. \yzcomment{we could go directly to results and analysis instead of explaining dataset once more since we have already used a large space in section 4.1 and 4.2 to introduce the two datasets.}
In Section~\ref{sec:result_lies}, we prove that our approach can outperform the LLM baseline and other baselines in predicting human lies. Section~\ref{sec:result_friction} shows that our approach has a high precision in predicting deception, potentially creating friction at just the point when a human player is making a critical decision. 

\subsection{Detecting Human Lies}
\label{sec:result_lies}
With human lie annotations, Table~\ref{tab:1K_results} compares \mbox{\textbf{CTRL-D}} prediction to these baselines: 1) \textbf{LLM baseline using Direct Judgment}, 2) \textbf{LLM baseline using Alignment Judgment} and 3) \textbf{Context LSTM with power} by \citet{peskov2020takes} 4) \textbf{Human suspected lies} by \citet{peskov2020takes}. This section answers these questions:
\begin{enumerate}
    \item Is our value-based RL with classifier approach (CTRL-D) able to predict human lies? 
    \item How good it is when compared to baselines?
    \item What lies does it miss (False negatives) and what truth does it misconceive as lies (False positives)?
\end{enumerate}

\paragraph{Our detection is most effective at predicting human lies.} It predicts with high precision \(0.950\), compared to both LLM-based methods and to Peskov's LSTM. In other words, when CTRL-D predicts a message as a lie, it is\fgcomment{?} 95\% likely that the message is an \textbf{actual} human lie. However, our detection is not perfect. Its recall is \(0.238\), meaning that it can only catch \textbf{one} out of every \textbf{four} lies. This high-precision but low-recall behavior is still valuable in deployment: since human players alone perform poorly at lie detection (precision = \(0.252\)), they can rely on CTRL-D’s predictions when it flags a message as a lie.


\paragraph{The LLM baseline is good at recall but bad at precision.} Comparing the recall of CTRL-D with our LLM baseline with Direct Judgment, the latter has good recall but low precision. The LLM predicts 45.5\% of all messages as lies, which is too sensitive to be useful, when considering the number of misaligned \cicero predicted orders and proposed orders seen in a message. Current LLM-based approaches to providing guidance do not appear capable of recognizing deception, and the resulting high false positive rate makes this strategy not compatible for real-time human games. 

\paragraph{CTRL-D misses deception if explicit actions are unclear in negotiations.} To characterize the lies CTRL-D misses detecting (False negatives), 41 of 61 non-detected lies have no logical forms for negotiations, while the remaining 20 have errors in logical forms when we parse the messages from English to \abr{amr}. 
We examine the remaining 20 and correct errors in logical forms by hand, to see if doing so improves prediction. With \textbf{logical forms annotated} by humans, the recall of CTRL-D (Table~\ref{tab:1K_results}) improves slightly from \(0.238\) to \(0.300\), showing that CTRL-D depends on proper logical parsing of natural language text. We further narrow the test dataset to include only messages that explicitly propose moves—allowing us to compute \textit{bait}, \textit{switch}, and \textit{edge}, the three features used to estimate deception. Under this setting, recall increases to \(0.413\), resulting in a higher F1-score compared to detecting deception across all messages. While CTRL-D performs best at identifying deceptive proposals involving explicit moves (via reinforcement learning), it also generalizes to detect lies in broader messages through text-based embeddings.
\jbgcomment{unclear}
\begin{table}[t]
    \begin{tabular}{p{1.2cm}p{5.5cm}}
    \hline
    \textbf{Sender} & \textbf{Message}   \\ 
    \hline
     \rowcolor{grayish} Germany (Truth) & \small{Well the feeling is mutual. I wanted to let you know that Austria asked for my help putting pressure on Warsaw. I don't intend to do that, but \textbf{I recommend you use Silesia to support your Rumanian unit into Galicia}. I promise not to interfere with this maneuver if you promise to keep your baltic fleet focused on defending Sweden from England?} \\ 
  \hline
    \end{tabular}
    \caption{An example of False Positive that  CTRL-D detects. Germany has a possible short-term gain had they betrayed Russia, but they nonetheless followed through on their proposal.
    }
    \label{tab:fp_cfrl}
\end{table}

\begin{table}[t]
    \begin{tabular}{p{1.2cm}p{5.5cm}}
    \hline
    \textbf{Sender} & \textbf{Message}   \\ 
    \hline
     \rowcolor{grayish} France (Truth) & \small{I am supporting Tys to Wes. Can you use Mar to support spain hold?} \\ 
  \hline
    \end{tabular}
    \caption{An example of False Positive that the LLM baseline detects as deception. It interprets France's message as a promise to support Austria's order. \cicero{}'s predict orders for France with \amr{A MUN H} where LLM baseline misinterprets as France is attacking Germany. It claims that France contradicts.}
    \label{tab:fp_llama3}
\end{table}

\paragraph{CTRL-D mispredicts once, while LLM baselines mispredict frequently.} We further investigate the false positives of CTRL-D and LLM baselines. Since our approach is very precise, it predicts only one true message as a lie (Table~\ref{tab:fp_cfrl}). On the other hand, the LLM baseline with Direct Judgment\footnote{We focus on Direct Judgment and omit the Alignment Judgment baseline variant since both LLM approaches are similar and have similar results.} mispredicts \(412\) messages as lies (Example in Table~\ref{tab:fp_llama3}). The LLM baseline is heavily constrained on \cicero{}'s predicted orders when it considers proposed orders in messages. This could be improved if LLM baseline can recognize that there are many plausible possibilities for players' orders, which are not necessarily deceptive. 

In sum, CTRL-D captures human lies best among all methods, including LLM-based methods. Though LLM baseline is better with semantics, it still lacks skills to interpret Diplomacy information in a way that would enable deception detection. With a strong agent, \cicero, predicting human lies using its RL value function makes detection possible. To further validate the quality of our deception detection, we evaluate both CTRL-D and LLM baseline on a larger data set that contains interactions between humans and \cicero.

\subsection{Awareness against Deception}
\label{sec:result_friction}
\begin{table*}[h]
    \centering
    \begin{tabular}{lcc}
        \hline
        \textbf{Model} & \textbf{Deceptive prediction rate} &\textbf{Precision} \\
        \hline
        LLM baseline using Direct Judgment & 0.413 & - \\
        LLM baseline using Alignment Judgment & 0.066 & 0.282  \\
        CTRL-D \textbf{(ours)} & 0.014 & \textbf{0.727} \\
        Human Actual Lie Rate & 0.050 & -\\
        \hline
    \end{tabular}
    \caption{Human verification supports \textbf{CTRL-D} as the stronger method with higher precision. However, \textbf{LLM baseline using Alignment Judgment} is able to detect some lies. \textbf{LLM baseline using Direct Judgment} detect almost half of messages as deception. A rate of messages that humans label as lies is included for comparison \cite{peskov2020takes}.}
    \label{tab:meta_results}
\end{table*}

This section, we evaluate CTRL-D and LLM baselines using the \amr{webdiplomacy.net} data set. Since this dataset lacks human deception annotations, we first let both models predict whether each message is deceptive, then verify the labeled predictions through human judgment. Human reviewers review with historical messages and final orders that sender and recipient submit through games. This information helps determine whether a sender deceives a recipient by comparing between 1) the sender's commitment in a proposal and 2) the sender's final orders. Although this verification is limited to deception that appears within explicit orders of the sender, this could serve as more evidence to verify performance of our approach and the LLM baseline. 
\paragraph{CTRL-D has precision than the LLM baseline, which overpredicts deception.} 
Our findings are consistent with those on the previous dataset, that CTRL-D is the strongest to predict deception (Table~\ref{tab:meta_results}). LLM baseline with Direct Judgment predict 41.3\% of all samples as deceptive, which is greatly higher than 5\% actual lie rate from humans~\citep{peskov2020takes}. This high rate makes human verification impractical. For LLM baseline with Alignment Judgment, its precision is \(0.282\) (only 1 in 4 flagged messages is a true lie). 
\paragraph{Errors in CTRL-D and LLM baselines showing thier weakness.} 
We cross-validate our CTRL-D with LLM baseline under Alignment Judgment to evaluate their ability to detect deceptive proposals. While both methods can correctly identify some lies, each may fail under different circumstances. We present several examples here:

\begin{itemize}
    \item Both models label it deceptive, and indeed it is a lie (Table \ref{tab:tp_both}).
    \item LLM baseline overlooks Russia's convoy promise, but CTRL-D detects the unfair exchange (Table~\ref{tab:tp_cfrl}).
    \item   CTRL-D misses one lie, while LLM baseline correctly spots it (Table~\ref{tab:tp_llama3}). 
\end{itemize}

Human verification supports CTRL-D as the stronger method; however, the LLM baseline can still catch some lies. We hope to further test these approaches with human players, thus introducing additional \textit{friction} in real negotiation settings.
\section{Related work}
\textbf{Deception in Human Behaviors.}
Research on deception highlights key behavioral and cognitive cues, such as micro-expressions and inconsistencies from mental strain~\cite{ekman2003, vrij2008}. Multimodal analyses integrating verbal and nonverbal signals have further enhanced detection accuracy~\cite{depaulo2003}. Linguistic cues linked to betrayal in the game Diplomacy offer insights~\cite{niculae-etal-2015-linguistic}. 
Moreover, computational models using language cues have shown promise in detecting deception in text, though evaluations have been limited to small datasets and specific scenarios~\cite{deception_algorithm, hazra-majumder-2024-tell}. Despite progress, the complex dynamics of deception in human behavior remain underexplored.

\textbf{\abr{ai} Deception in Texts.}
With the rise of \abr{ai}-generated content, detecting textual deception is crucial. Linguistic and psycholinguistic analysis aids detection, while transformer models improve accuracy~\cite{ott2011, 10.1002/pra2.2015.145052010082}.
Prior work focuses on detecting \abr{ai} deception using external and internal methods~\cite{park2024ai}. External techniques like ``\textit{consistency checks}''~\cite{fluri2024evaluating} analyze \abr{ai} behavior for inconsistencies, while internal methods examine embeddings to detect dishonesty~\cite{azaria-mitchell-2023-internal, burns2024discoveringlatentknowledgelanguage}. 


\section{Conclusion and Future Work}
Our study confirms that with a well-trained value function, we can estimate deception signs---\emph{bait}, \emph{switch}, \emph{edge}---to predict deception. 
CTRL-D, our counterfactual RL against deceptive proposals, has a good recall and almost perfect precision, which can be helpful for humans that struggle to recognize deception within Diplomacy.
%
Comparing to an LLM baseline that is too sensitive with a higher recall, CTRL-D predicts human lies and generalizes, demonstrating high precision consistently on both evaluation data sets. 

While these tasks for deception detection are for Diplomacy, they illustrate the general risks and challenges of \abr{ai}-deception. 
Future human-\abr{ai} interaction tools can build on our methods to reevaluate trust in suspicious negotiations.
Applying CTRL-D to real-world deception remains challenging, particularly in open-ended contexts such as \textit{scams}, where individuals are manipulated into taking actions that compromise their financial security. One possible direction is to extend our method to more controlled environments involving less explicit actions, such as \textit{Avalon}~\citep{serrino2019friendfoe}, where deception occurs through proposing team subsets, and eventually to games like \textit{Among Us} and \textit{Werewolves}~\citep{jin2024learning,xu2025learningstrategiclanguageagents}, provided there is access to grounded actions, dialogue, and action-value estimations.
Looking further ahead, in more complex environments where reinforcement learning has shown strong performance, such as \textit{StarCraft} and \textit{Go}~\citep{Silver2016MasteringTG, Vinyals2019GrandmasterLI}, detecting deception through moves, combat actions, and in-game communication (e.g., language used during battles) presents an exciting direction for future research.

\section*{Limitation}
This study can evaluate through real-time Diplomacy games to test whether our approach could help trigger friction in human players and if it could, how useful it is. We limit our evaluation space to Diplomacy, and we could gain a better understanding of deception if we expand to broader areas like negotiation in trading. Our approach, \textbf{CTRL-D}, relies on a tool (\abr{amr}) to transform texts into logical forms. Its representation could sometimes be invalid and undermine accuracy deception detection. Our detection can only predict those negotiations with explicit actions, missing opportunities where deception occurs in other forms.

\section*{Ethical Considerations}
Our study uses existing data sets so we do not experiment or collect new data from humans. This paper highlights deception detection which will be necessary for dealing with existing and future harms of AI and LLMs.  As a double-edged sword, acknowledging this deception may make future systems better at masking their deception.

\section*{Acknowledgments}
We thank Meta for granting access to over 40,000 games played on the online platform \url{webdiplomacy.net} and for open sourcing Llama 3 and \cicero{}. 
Our thanks also go to Ulf Hermjakob and Tess Wood for Diplomacy \abr{amr} Annotation Dictionary.
We thank Alex Hedges for LLM Diplomacy setup and Sadra Sabouri for valuable feedback.
This research is supported by the U.S. Defense Advanced Research Projects Agency (DARPA) Other Transaction award HR00112490374 from the Friction for Accountability in Conversational Transactions (FACT) program.
Any opinions, findings, conclusions, or recommendations expressed here are those of the authors and
do not necessarily reflect the view of the sponsors.

\bibliography{bib/custom}




\appendix
\renewcommand{\thefigure}{A\arabic{figure}}
\setcounter{figure}{0}
\renewcommand{\thetable}{A\arabic{table}}
\setcounter{table}{0}


\section{Lie Categories}
\label{sec:denis_lie_cat}

We simply categorize lie messages using keywords, prioritize as the following:
\begin{itemize}
    \item \textbf{Deceptive Moves.} \{support, move, attack, retreat, convoy, hold, bounce\} 
    \item \textbf{Feigning Trust/Loyalty.} \{trust, friend\}
    \item \textbf{Withholding Information.} \{no idea, not sure\}
    \item \textbf{False Excuse.} \{sorry, busy\}.
\end{itemize} 
If the message does not belong to any category, we rule it as \textbf{Other}. Examples for each category in Table \ref{tab:lie_category}.

\section{Classifier Discussion}
\begin{table}[h]
\centering
\begin{tabular}{lcc}
\hline
\textbf{} & \textbf{Precision} & \textbf{Recall} \\
\hline
Rule-based & 0.140 & 0.429 \\
Linear sum with weights & 0.250 & 0.214 \\
CTRL-D (\textbf{ours}) & \textbf{0.950} & 0.238 \\
\hline
\end{tabular}
\caption{Comparison of different deception detection methods.}
\label{tab:classifier}
\end{table}
As there is a concern that a BERT-based classifier may learn a simple heuristic, for example, flagging offers that seem ``\textit{too good to be true}'', rather than truly understanding deception. First, we want to provide evidence that the classifier does not heuristically predict deception. In our preliminary experiments, we first attempted the approach using heuristics directly to deceptive values: 
\begin{itemize}
    \item Rule-based: $U_1 > T_1, U_2 > T_2, U_3 > T_3$
    \item Linear sum with weights and a threshold: $U_1w_1 + U_2w_2+ U_3w_3 > T$
\end{itemize}
where results were not good when comparing to the CTRL-D classifier (Table~\ref{tab:classifier}).

\begin{table}
\centering
\begin{tabular}{lccc}
\hline
\textbf{} & \textbf{Precision} & \textbf{Recall} & \textbf{F1-score} \\
\hline
BERT (\textbf{ours}) & \textbf{0.950} & 0.238 & 0.380 \\
RoBERTa (new) & 0.767 & \textbf{0.413} & \textbf{0.537} \\
\hline
\end{tabular}
\caption{Performance comparison of CTRL-D trained on BERT vs RoBERTa.}
\label{tab:bert_comparison}
\end{table}

With a stronger text embedding, e.g. RoBERTa~\citep{Liu2019RoBERTaAR}, we further run additional experiment that replacing BERT with RoBERTa for text embedding and keep the rest of models. The results (Table~\ref{tab:bert_comparison}) show RoBERTa can catch more lies. However, we are unsure why precision is slightly lower.

\section{Additional Llama experiments}

We apply Chain-of-Thought~\citep[\abr{CoT,}][]{wei2022cot} to our LLM baselines. To simplify this, we directly applied CoT into \textit{Alignment Judgement}, which is one of our LLM baselines that compares the \cicero's expected action and a dialogue of a certain player to find misalignment that could be a deception. As in Table \ref{tab:llm_add_results}, adding a chain-of-thought by explicitly prompting a text ``\textit{step-by-step}'' did not improve from a baseline, Alignment Judgment. We hypothesize that 
\begin{enumerate}
    \item it is still challenging for LLM to recognize  patterns from deviations of \cicero's predicted moves and moves that appear in a conversation in natural language and
    \item a possibility of next moves are too wide and \cicero's predicted moves could not represent a real human intent, relying solely on predicted moves could be impossible. 
\end{enumerate}

Another experiment with LLM is providing deceptive values (\textit{bait}, \textit{switch} and \textit{edge}) to our LLM baseline, which results in an LLM triggering lies even more often that its \textbf{recall} is the highest among all baselines; however, the precision of detecting is worsened. Overall, these additional experiments on LLM concretely prove that LLM struggles to detect lies.
\begin{table}[h]
    \begin{tabular}{p{1.2cm}p{5.5cm}}
    \hline
    \textbf{Sender} & \textbf{Message}   \\ 
    \hline
     \rowcolor{grayish} Austria (Lie) & \small{Germany is apparently ripe for the taking. I'll support you into Kiel in exchange for support into Bulgaria from Serbia.} \\ 
  \hline
    \end{tabular}
    \caption{LLM baseline and our CTRL-D correctly detect this messages as a lie, where Austria does not support to Kiel as they propose.}
    \label{tab:tp_both}
\end{table}

\begin{table}[h]
    \begin{tabular}{p{1.2cm}p{5.5cm}}
    \hline
    \textbf{Sender} & \textbf{Message}   \\ 
    \hline
     \rowcolor{grayish} Russia (Lie) & \small{Oh and RE: \textbf{St Petersburg yeah. My plan is to convoy an army over from Norway.} \textbf{If you vacate Serbia} I would be happy to then move my units back to Ukraine and Sevastopol in return for you then moving that Fleet to Smyrna.} \\ 
  \hline
    \end{tabular}
    \caption{Russia lies to Austria which CTRL-D is able to catch this, while LLM baseline is not.}
    \label{tab:tp_cfrl}
\end{table}

\begin{table}[h]
    \begin{tabular}{p{1.2cm}p{5.5cm}}
    \hline
    \textbf{Sender} & \textbf{Message}   \\ 
    \hline
     \rowcolor{grayish} Germany (Lie) & \small{Hi again. \textbf{My new army in kiel will move to munich and Munich will move to bohemia.} You are a trustful neighbor and we can share victory. Im not moving North sea and i do not expect you to do it. Let the Russian collapse by himself} \\ 
  \hline
    \end{tabular}
    \caption{LLM baseline detects this lie from Germany. However, CTRL-D misses to detect this lie due to its error in logical forms.}
    \label{tab:tp_llama3}
\end{table}
\begin{table*}[h]
    \centering
    \begin{tabular}{lccc}
        \hline
        \textbf{Model} & \textbf{Precision} & \textbf{Recall} & \textbf{F1-Score} \\
        \hline
        LLM baseline: Direct Judgment & 0.095 & 0.551 &0.161 \\
        LLM baseline: Alignment Judgment & 0.147 & 0.065 & 0.090  \\
        LLM baseline: Alignment Judgment and CoT & 0.094 & 0.288 &0.141 \\
        LLM baseline: bait, switch and edge values & 0.081 & \textbf{0.575} & 0.142 \\
        CTRL-D \textbf{(ours)} & \textbf{0.950} & 0.238 & \textbf{0.380} \\
        \hline
    \end{tabular}
    \caption{Additional experiments on prompting Llama with Chain-of-Thought and three deceptive values from our bait, switch and edge estimation.}
    \label{tab:llm_add_results}
\end{table*}

\begin{table*}[t]
    \begin{tabular}{p{11cm}p{4cm}}
    \hline
    \textbf{Message} & \textbf{Category}   \\ 
    \hline
    \small{We have a terrific counter, if you’d like to work with me. 

I can tap Greece and Aegean, which should allow you to save Bulgaria (Const S Bulgaria holding). At the same time, Black Sea Support Armenia to Sevastopol WILL WORK! I’m planning to move over to Trieste, so the end result of all this would be Turkey regaining Sev and remaining at 5 while I take Trieste and get to 5. From there, I think we should be able to work together to finish off Austria.} & \small{Deceptive Moves} \\ 
    \rowcolor{grayish}  
    \small{This is too big a risk for me. My preferred move is to Belgium. I understand this may warrant doing a couple of favors for your next year in return. But I just feel too exposed otherwise. I'm about to head out for the evening and I have my move set for Belgium. I'm hoping for your support into there, but I understand if you can't provide it.} & \small{Deceptive Moves} \\ 
    \small{I can move on Picardy, but only if you move on Belgium.  I think Germany is going for him, so now’s your chance to get a slice of France while the getting is good.} & \small{Deceptive Moves} \\ 
    \rowcolor{grayish}  
    \small{Hi Turkey! I’m sorry that I’ve been so slow to get in touch. Kind of a rough day for me to begin a game as I e been pretty swamped. Things are clearing up now, and I appreciate you reaching out to me. 

    So far I have notes from Austria and Russia being pretty cagey and non-committal. Perhaps that is just the life of Italy? Nobody really has me in their plans?
    
    I don’t really know what I’m going to do yet, so if you have ideas, or you have a use for me, please let me know. I’d basically be delighted to work with anyone who really wants to work with me. (No sign yet that this includes anyone at all)} & \small{False Excuse} \\ 
    \small{I am a bit new at this stuff, sorry} & \small{False Excuse} \\ 
    \rowcolor{grayish} \small{But yeah sorry about that, I had put that in as my orders and then forgot to change them} & \small{False Excuse} \\ 
    \small{If you want me to trust you I think you should give at least one back, as you are much bigger and frankly it's probably in your best interest to go for the solo, although if the rest of the contries cooperate you almost certaintly won't get it. And your definitly going to need me if you want to go for the draw} & \small{Feigning Trust/Loyalty} \\
    \rowcolor{grayish} \small{Well good luck to you too! Let's be friend! Have you heard anything interesting?} & \small{Feigning Trust/Loyalty}\\
    \small{I can't say ik certain of how things will go in the west but I like to maintain peace as long as possible and a few allies I can trust is important as hell in this game} & \small{Feigning Trust/Loyalty}\\
    \rowcolor{grayish} \small{Not sure what your plan is. I'm covering Marseilles, Paris, and Belgium so you have no chance to take a supply center this turn.} & \small{Withholding Information}\\
    \rowcolor{grayish} \small{I’m not sure yet. Lots to think about.} & \small{Withholding Information}\\
    \rowcolor{grayish} \small{How do you know they are plotting each-others destruction? Not sure which side to take yet.} & \small{Withholding Information}\\
    \small{Hmm I thought it was something else} & \small{Other}\\
    \rowcolor{grayish} \small{I can settle for survival. Deal.} & \small{Other}\\
    \small{I can't share screenshots apparently, but that's true.  It's not just England} & \small{Other}\\
    \end{tabular}
    \caption{Our lie categories using examples from \citet{peskov2020takes}}
    \label{tab:lie_category}
\end{table*}
\onecolumn
\begin{figure}[ht]
\centering
\fbox {
    \parbox{\linewidth}{
You are an expert assistant specializing in the Diplomacy board game. Your role is to assist a novice player by analyzing:  

        1. The current board state.  
        
        2. The message history exchanged between the novice player and the counterplayer.  
        
        3. The predicted orders for the counterplayer.  

        Your primary objective is to evaluate whether the counterplayer's predicted orders align with the message history and the board state. 

        Key Evaluation Guidelines:

        1. Consider an order aligned if its purpose or intent is consistent with the counterplayer's stated goals or the tactical/strategic needs implied by the board state.
        
        2. Special orders (e.g., supporting moves from counterplayer) can still be aligned if it serves to ensure the success of a critical move or maintains flexibility in uncertain situations.
        
        3. Misalignment occurs if the order: 
            Contradicts the stated strategy or creates unnecessary risks. 
            Fails to contribute meaningfully to the position given the board state.
            
        4. Evaluate the broader context of each move, including defensive, offensive, or preparatory purposes, and weigh these factors in your analysis. For each predicted order, explain whether it aligns with the message history and board state. Use reasoning that incorporates tactical, strategic, and contextual factors while avoiding overly strict judgments. Limit your evaluation to alignment or misalignment, with short reasoning for each.

        **Predicted Orders of ITALY:**  
        
        ['A TUN - ALB VIA', 'F ION C A TUN - ALB', 'F NAP - TYS', 'A VEN - TRI']

        **Board State:**  
        
        AUSTRIA: ['F TRI', 'A SER', 'A GAL', 'A BUD'], ENGLAND: ['F LON', 'F NWY', 'A YOR', 'F EDI'], FRANCE: ['A POR', 'A SPA', 'F BEL', 'F BRE', 'F MAR', 'A PAR'], GERMANY: ['F DEN', 'A MUN', 'A HOL', 'A BER', 'F KIE'], ITALY: ['A VEN', 'F ION', 'A TUN', 'F NAP'], RUSSIA: ['A WAR', 'F SEV', 'F BOT', 'A RUM', 'A STP'], TURKEY: ['A CON', 'A BUL', 'F BLA', 'F SMY']

        **Message History:**  
        
        Message from AUSTRIA: "France and Turkey are trying to get me to attack you, obviously I will not do that."
        
        Message from ITALY: "With the build in Marseilles, France's goals are quite transparent."
        
        Message from ITALY: "Turkey will likely bounce me in EAS. Would you be willing to support me into Greece (via convoy)?"
        
        Message from ITALY: "From there, I can support you into Bulgaria."
        
        Message from AUSTRIA: "I'm willing to debate it but I had a different view on this turn. I figured he would bounce you in EAS but what if you went for AEG instead? Unless you're worried about overextending yourself due to France. Because I trust you enough at this point to move Tri to Alb and then we will definitely be able to get Gre in the fall."
        
        Message from ITALY: "Which makes going to AEG/EAS a bit of an overstretch."
        
        Message from ITALY: "If we can get Germany on board I'd feel a lot more comfortable with committing east."
        
        Message from ITALY: "Sorry, some connection issues on my end so messages sent out of order."

        **Question(Direct Judgment):**  
        
        I am playing as AUSTRIA. You need to determine whether ITALY is being deceptive based on the above information. Your answer must be one of two categories: Truthful or Deceptive. Provide a brief explanation and highlight any potentially deceptive content.

        **Question(Alignment Judgment):**  
        
        I am playing as AUSTRIA. For each of ITALY's predicted orders, evaluate whether it aligns with the message history and the current board state. Explain the orders first and provide short reasoning and analysis for each predicted move.
    }
}
\caption{Prompt template used in Llama baselines for identifying alignment and detecting deception.}
\label{fig:prompt_llama}
\end{figure}

\end{document}